\documentclass[journal]{IEEEtran} 

\usepackage{cite}
\usepackage{booktabs}
\usepackage[pdftex]{graphicx}
\usepackage{amsmath}
\usepackage{url}
\usepackage{hyperref}
\usepackage{color}
\usepackage{multirow}
\usepackage{cite}

\usepackage[caption=false,font=footnotesize]{subfig}

\begin{document}
%
\title{Spatio-Temporal Vegetation Pixel Classification \\By Using Convolutional Networks}
%
%
%

\author{Keiller~Nogueira,
	Jefersson~A.~dos~Santos,
	Nathalia~Menini,
	Thiago~S.~F.~Silva, \\
	Leonor~Patricia~C.~Morellato,
	and Ricardo~da~S.~Torres
\thanks{K.~Nogueira, and J.~A.~dos~Santos are with the Department of Computer Science, Universidade Federal de Minas Gerais (UFMG), Brazil. Emails: \{keiller.nogueira, jefersson\}@dcc.ufmg.br
\newline Nathalia Menini and Ricardo da S. Torres are with Institute of Computing, University of Campinas, Brazil. 
\newline L.~P.~C.~Morellato is with Instituto de Biociências (IB), Universidade Estadual Paulista (Unesp), Brazil.
\newline T.~S.~F.~Silva is with IGCE/Unesp, Brazil and with Biological and Environmental Sciences, Faculty of Natural Sciences, University of Stirling, UK.}%
\thanks{The authors thank the Pr{\'o}-Reitoria de Pesquisa da UFMG; FAPEMIG (APQ-00449-17); S{\~a}o Paulo Research Foundation FAPESP (2013/50155-0, 2013/50169-1, 2009/54208-6, 2016/26170-8, 2018/06918-3); CNPq (424700/2018-2); CNPq Research fellowship to JAS, LPCM, RST, TSFS; Coordenação de Aperfeiçoamento de Pessoal de Nível Superior - Brasil (CAPES) - Finance Code 001 (88881.145912/2017-01); the Cedro T{\^e}xtil, Reserva Vellozia, Parque Nacional da Serra do Cip{\'o}, PELD-CRSC-17.}}

\maketitle

\begin{abstract}
	Plant phenology studies rely on long-term monitoring of life cycles of plants. High-resolution unmanned aerial vehicles (UAVs) and near-surface technologies have been used for plant monitoring, demanding the creation of methods capable of locating and identifying plant species through time and space.
	However, this is a challenging task given the high volume of data, the constant data missing from temporal dataset, the heterogeneity of temporal profiles, the variety of plant visual patterns, and the unclear definition of individuals' boundaries in plant communities.
	In this letter, we propose a novel method, suitable for phenological monitoring, based on Convolutional Networks (ConvNets) to perform spatio-temporal vegetation pixel-classification on high resolution images.
	We conducted a systematic evaluation using high-resolution vegetation image datasets associated with the Brazilian Cerrado biome.
	Experimental results show that the proposed approach is effective, overcoming other spatio-temporal pixel-classification strategies.
\end{abstract}


%
\IEEEpeerreviewmaketitle

\section{Introduction} \label{sec:intro}

Plant phenology is the science dedicated to the study of the life cycles of plants and their relations with climate conditions.
One of the most important challenges faced by {\em remote} phenology studies refers to the location (identification) of plant individuals of interest within images. Specially, in tropical regions, plant individuals of a large number of species co-exist {\em spatially} (community level) over time. Also, as species phenological traits for different species often differ from each other, species {\em temporal} profiles are expected to encode relevant discriminative properties. Properly combining spatial and temporal cues are, therefore, of paramount importance for the creation of effective vegetation pixel-classification methods. 

Spatio-temporal vegetation pixel classification, however, is a very challenging task~\cite{Anderson2013}, as it requires to handle:
(i) high volumes of temporal data;
(ii) missing data~\cite{garciaPatternMissingData2010}, which is a common aspect of temporal datasets;
(iii) heterogeneity of temporal patterns, due to differences between seasons, which impact on, for instance, the leafing and flowering of plants;
(iv) variety of plant patterns, given the high intraclass variance and high interclass similarity of species; and 
(v) unclear definition of individuals' boundaries, as canopies often overlap.

Several studies have been addressing those challenges~\cite{Almeida2016PRL,di2017end,kussul2017deep,song2018spatiotemporal}.
Some of them~\cite{Almeida2014EcoInfo,Almeida2016PRL,schafer2018classifying} employ general-purpose hand-crafted descriptors to represent image regions. Those methods are data dependent, requiring a full set of experiments to determine the most suitable descriptors for each application~\cite{keiller2016icpr}.
Other initiatives~\cite{di2017end,kussul2017deep,song2018spatiotemporal} rely on deep learning approaches~\cite{deeplearningbook}, such as ConvNets (or CNN), which are capable of learning, at once, data-driven features and classifiers.
However, those studies~\cite{di2017end,kussul2017deep,schafer2018classifying,song2018spatiotemporal} often focus on spatio-temporal pixel classification of land-use datasets.
To the best of our knowledge, none of those deep learning-based initiatives have been exploiting spatio-temporal properties of high-resolution vegetation images in pixel classification problems.


In this paper, we propose a deep-learning-based technique that fills this gap.
In other words, we propose a novel ConvNet for vegetation pixel classification in high-resolution temporal images acquired by UAVs and near-surface digital cameras.
The method can effectively learn a combined representation of images using distinct spatio-temporal properties.
Specifically, the proposed network has initial branches, which are responsible for learning spatial patterns from images of a specific timestamp.
All branches are unified into a unique network, which is in charge of combining information from previously learned spatial properties, creating a new spatio-temporal representation.
Since the whole process is integrated into a single framework that receives several patches (one for each timestamp) and outputs a single class label, the network can be trained end-to-end using the well-known backpropagation algorithm~\cite{deeplearningbook}.
Furthermore, since each branch handles a specific timestamp, the proposed approach can be easily adapted to handle missing data~\cite{garciaPatternMissingData2010} by using dropout~\cite{srivastava2014dropout}.


\section{Multi-Temporal ConvNet} \label{sec:method}


The proposed approach, called Multi-Temporal ConvNet, is based on pixelwise classification~\cite{keiller2016icpr}.
In this technique, each pixel of the image is independently processed and classified by the network.
Since the pixel itself has not enough information to be extracted from the network, it is usually represented by a context window, which aggregates spatial information to the learning process.
Technically, a \textit{context window} is a fixed-size patch with the pixel that the method should classify centered on it.
Overall, the context windows, generated for each and every pixel of the image, are delivered to the ConvNet, which processes them as standard images, outputting a unique classification which is related to the centered pixel of the currently processed window.

Networks proposed for pixelwise classification aggregate only spatial information.
Our vegetation pixel-classification method proposes to incorporate temporal information into pixelwise classification networks in order to learn a spatio-temporal model.
Precisely, instead of receiving only one context window, the proposed model receives as input several context windows, all of them centered on the same pixel along the temporal domain.
These windows are processed by several branches, each one related to an entry in the temporal domain.
Each branch receives a single context window that is directly connected to the \textit{timestamp}, i.e., there is an exact match between the branch and context window (since both should be related to the same timestamp).
These branches are directly responsible for learning spatial patterns related to the respective \textit{timestamp}.
All these branches are then depth-wise concatenated and processed by further layers of the network, responsible for combining the previously learned information by extracting patterns related to the whole time series and creating a spatio-temporal representation.
This final representation is, then, used to classify the pixel centered on the context windows.
This whole process is integrated into a unique network that receives several patches and outputs a class label, allowing the ConvNet to be trained end-to-end using backpropagation~\cite{deeplearningbook}.

\subsection{Network Architecture} \label{subsec:arc}

\begin{figure}[t!]
	\centering
	\includegraphics[width=\columnwidth]{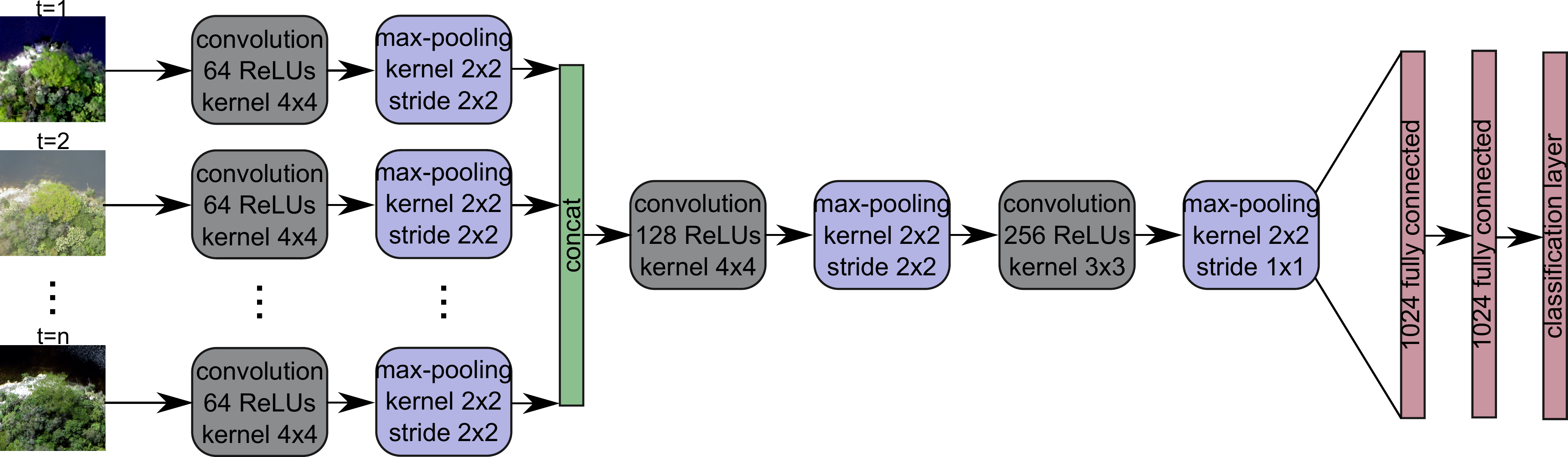}
	\caption{ConvNet architecture for spatio-temporal pixel classification.}
	\label{fig:net}
\end{figure}

The multi-temporal ConvNet architecture, proposed to perform spatio-temporal pixel-classification of vegetation images, is presented in Figure~\ref{fig:net}.
This network receives as input $25\times25$ context windows.
These patches (one for each timestamp) are processed by their respective network branch, which is responsible for capturing the spatial information of that specific timestamp.
All of these branches have a convolution (with 64 neurons and kernel of size $4\times4$) and a max-pooling operation (with both kernel and stride of size $2\times2$).
Then, all branches are depth-wise concatenated and further processed by two other layers, both composed of convolution and max-pooling operations.
In these layers, the convolutions have 128 (with $4\times4$ kernels) and 256 (with $3\times3$ filters) neurons, respectively.
Both max-pooling operations use the same kernel ($2\times2$) but differ in the stride: the first one uses $2\times2$ while the second employs $1\times1$.
Finally, fully connected layers create the final classification of the pixel centered on the input context windows.
Rectified Linear Unit (ReLU)~\cite{deeplearningbook} was the processing units used in all layers.
Our implementation also includes:
(i) batch normalization~\cite{deeplearningbook}, which is employed after each convolution, and
(ii) dropout~\cite{srivastava2014dropout}, which is mainly used within the fully connected layers to avoid overfitting.

\subsection{Missing Data Adaptation} \label{subsec:missing_data}

Since the proposed method has a branch for each timestamp, it can be easily adapted to handle one of the challenging problems of temporal series: missing data~\cite{Anderson2013}.
Very common in spatio-temporal datasets, this issue is defined as the lack of stored information for a timestamp of the time series.
This problem can have a significant effect on the conclusions that can be drawn from the data~\cite{garciaPatternMissingData2010}.
In order to handle this crucial problem, we propose to use dropout~\cite{srivastava2014dropout}.

Dropout~\cite{srivastava2014dropout} is a regularization technique that drops some neurons during the \textbf{learning} process.
Technically, for each learning iteration of the network, dropout method randomly selects (based on a predefined probability, which is usually 50\%) a set of neurons that: 
(i) do not contribute (i.e., propagates zero) during the forward step, and 
(ii) do not receive updates during the backpropagation phase.
Therefore, this set of neurons does not participate in the learning process during an iteration.
This allows the creation of an ensemble of models that share parameters, improving the representation learned by the network.
During the \textbf{inference} step, all neurons are preserved and contribute to the final outcome of the network.

In order to handle missing data, we proposed several small changes in the method and in the dropout technique.
One change is the inclusion of a dropout layer just before the concatenation layer, which is responsible for dropping some neurons (or, in a broader view, branches).
This new layer acts in its traditional way during the learning phase with the probability (to randomly select the neurons that are preserved) set to $1/t$, where $t$ is the temporal length.
This value is defined based on the worst scenario of missing data, i.e., when there is only 1 available timestamp in the whole temporal domain $t$.
Hence, by setting this probability following this idea, we force the network to learn a model that should extract all feasible information from the available data without being hampered by the missing data (even if it is the worst scenario).
Although working traditionally during the learning phase, dropout layer works differently during the prediction step.
Instead of preserving all neurons, the dropout layer drops the branches (or timestamps, since each branch is correlated to a timestamp) with missing data, which do not contribute to the final result.


\section{Experimental Setup} \label{sec:protocol}


\subsection{Phenological Temporal Datasets} \label{subsec:dataset}


\subsubsection{Itirapina Dataset} \label{subsubsec:itarapina}

The Itirapina dataset~\cite{Almeida2016PRL} comprises images collected with a near-remote phenological system composed of a camera set up in an 18m tower in a Cerrado sensu stricto, a savanna-like vegetation located at Itirapina, S{\~a}o Paulo State, Brazil.
This camera was set up to automatically take hourly photos (at $1280\times960$ pixels of resolution) from 6:00 to 18:00 h (UTC-3) between August 29th and October 3rd, 2011 (resulting in \textbf{36 days or timestamps}).
These images were then labeled by experts as belonging to one of the six possible plant species, as presented in Table~\ref{tab:itarapina}.
Example of a timestamp and the ground truth are presented in Figure~\ref{fig:itarapina}.

\newcommand{\exFigSize}{0.17}

\begin{table} 
	\begin{minipage}{0.5\linewidth} 
	\caption{Itirapina dataset} 
	\label{tab:itarapina} 
	\centering 
	\resizebox{0.9\columnwidth}{!}{ 
	\begin{tabular}{@{}lr@{}}
		\toprule
		\textbf{Class}                  & \multicolumn{1}{l}{\textbf{\#pixels}} \\ \midrule
		\textbf{\textit{Aspidosperma tomentosum}}           		& 4,040                                       \\
		\textbf{\textit{Caryocar brasiliensis}}  					& 6,601                                      \\
		\textbf{\textit{Myrcia guianesis}} 							& 2,630                                      \\
		\textbf{\textit{Miconia rubiginosa}}    					& 6,715                                      \\
		\textbf{\textit{Pouteria ramiflora}}    					& 2,373                                      \\ 
		\textbf{\textit{Pouteria torta}}    						& 2,037                                      \\ 
		\bottomrule
	\end{tabular}}
	\end{minipage}%
	\begin{minipage}{0.5\linewidth} 
		\caption{Serra do Cip{\'o} dataset.}
		\label{tab:serracipo}
		\centering 
		\resizebox{0.9\columnwidth}{!}{ 
		\begin{tabular}{@{}lr@{}}
			\toprule
			\textbf{Class}                  & \multicolumn{1}{l}{\textbf{\#pixels}} \\ \midrule
			\textbf{\textit{Vochysia cinnamomea}}           		& 34,754                                       \\
			\textbf{\textit{Eremanthus erythropappus}}  			& 31,250                                      \\
			\textbf{\textit{Bowdichia virgilioides}} 				& 33,137                                      \\
			\textbf{\textit{Set of evergreen species}}    	& 48,041                                      \\ 
			\bottomrule
		\end{tabular}}
	\end{minipage}%
\end{table}


\begin{figure}[!t]
	\centering
	\subfloat[18h - October 3rd, 2011]{
		\includegraphics[height=\exFigSize\textwidth]{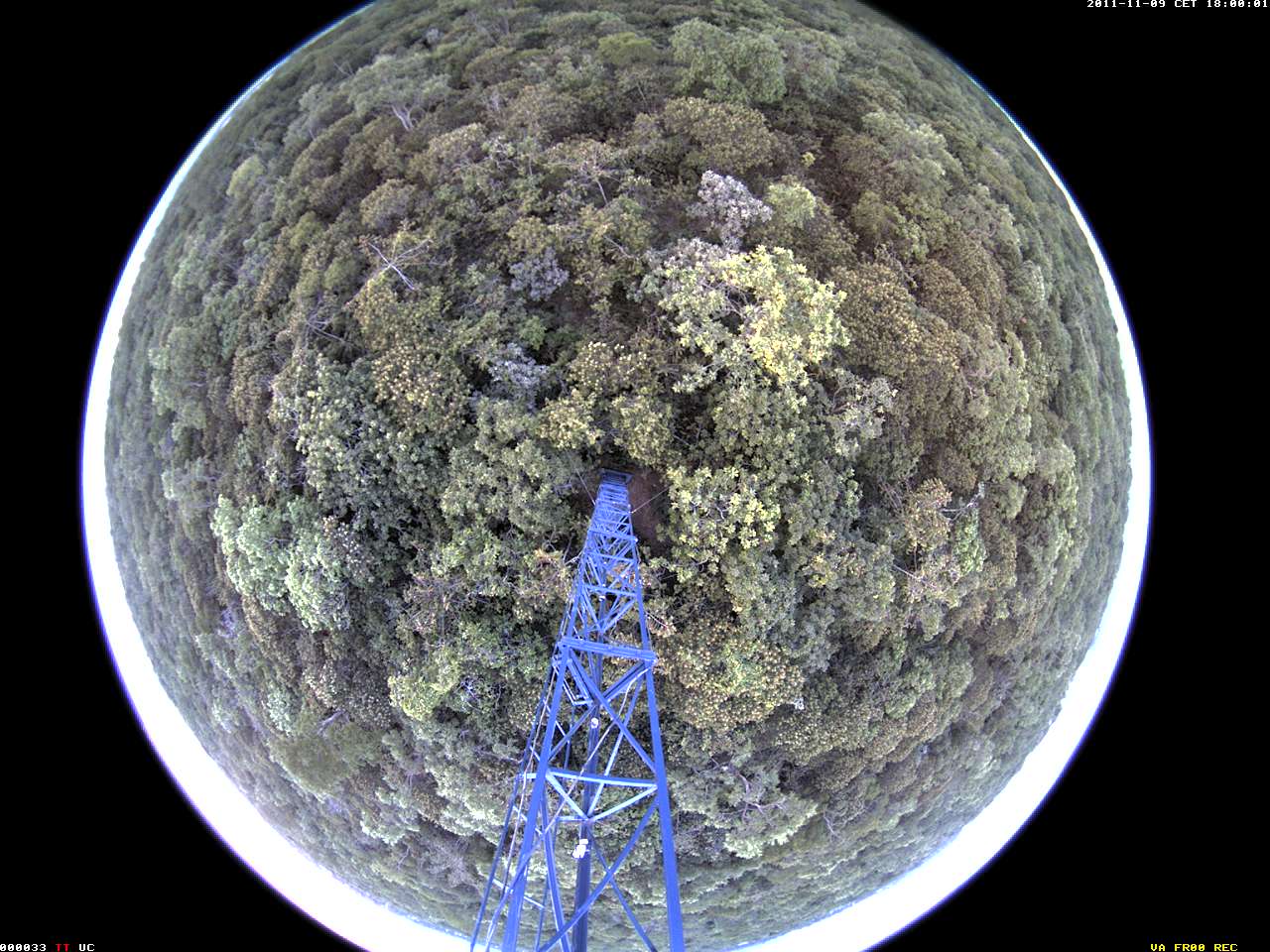}
	}
	\subfloat[Ground-Truth]{
		\includegraphics[height=\exFigSize\textwidth]{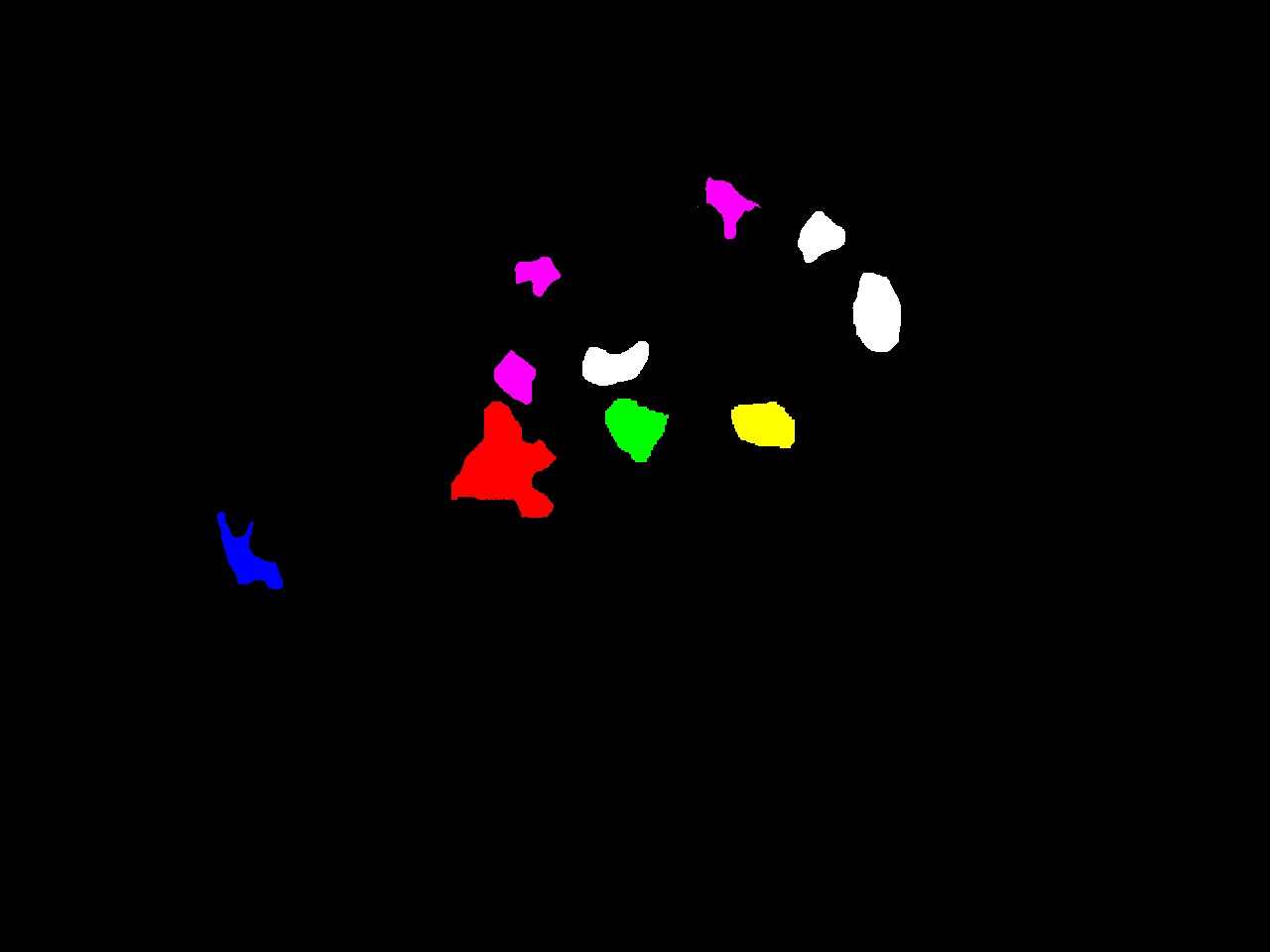}
	}
	\caption{Image of the Itirapina dataset and the ground truth. 
		Legend -- Black: unclassified/background. White: {\em Miconia rubiginosa}. Red: {\em Caryocar brasiliensis}. Green: {\em Myrcia guianesis}. Pink: {\em Aspidosperma tomentosum}. Blue: {\em Pouteria torta}. Yellow: {\em Pouteria ramiflora}.
	}
	\label{fig:itarapina}
\end{figure}

\subsubsection{Serra do Cip{\'o} Dataset} \label{subsubsec:serracipo}

\begin{figure}[!t]
	\centering
	\subfloat[December 24th, 2016]{
		\includegraphics[height=\exFigSize\textwidth]{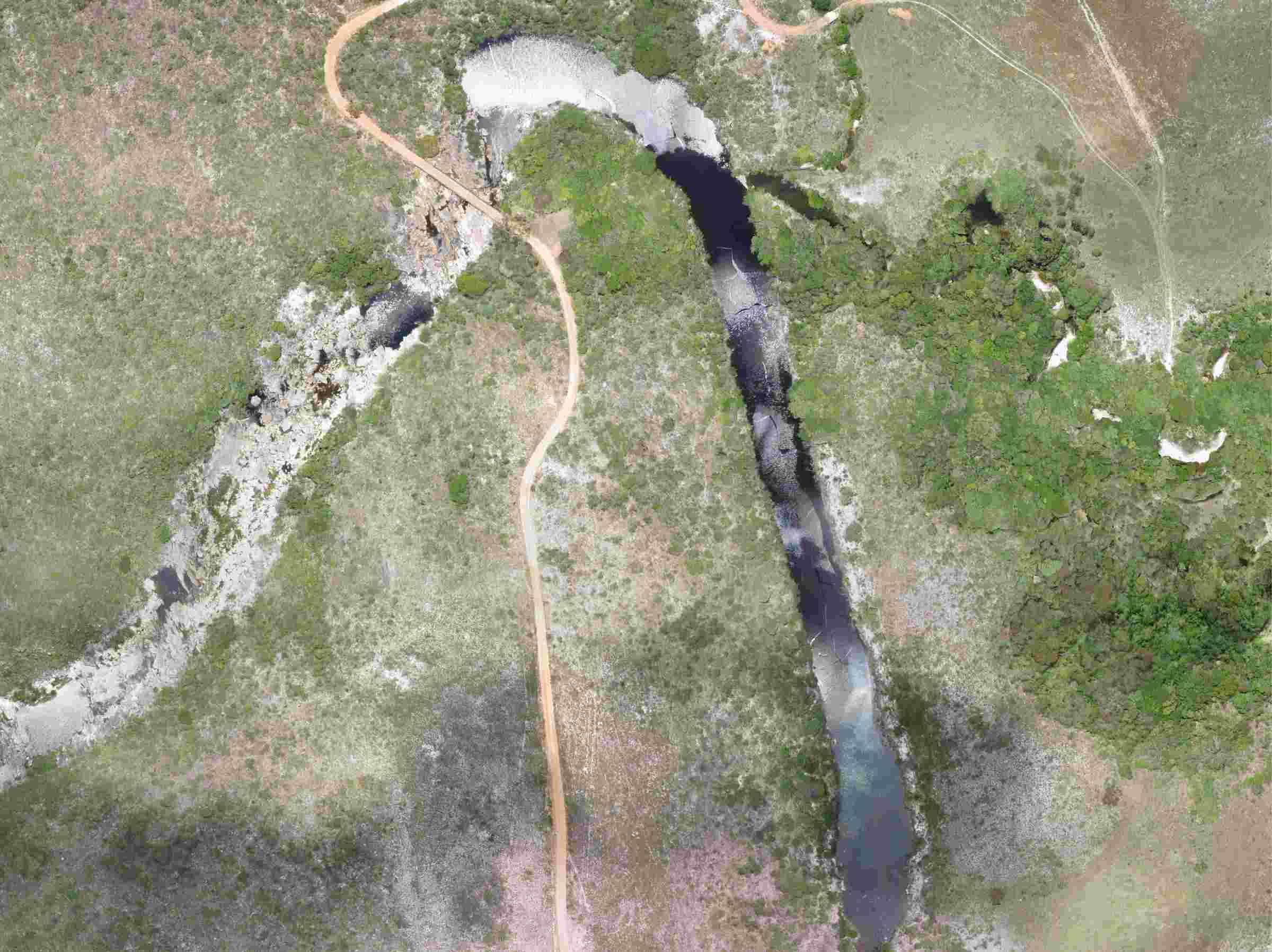}
	}
	\subfloat[Ground-Truth]{
		\includegraphics[height=\exFigSize\textwidth]{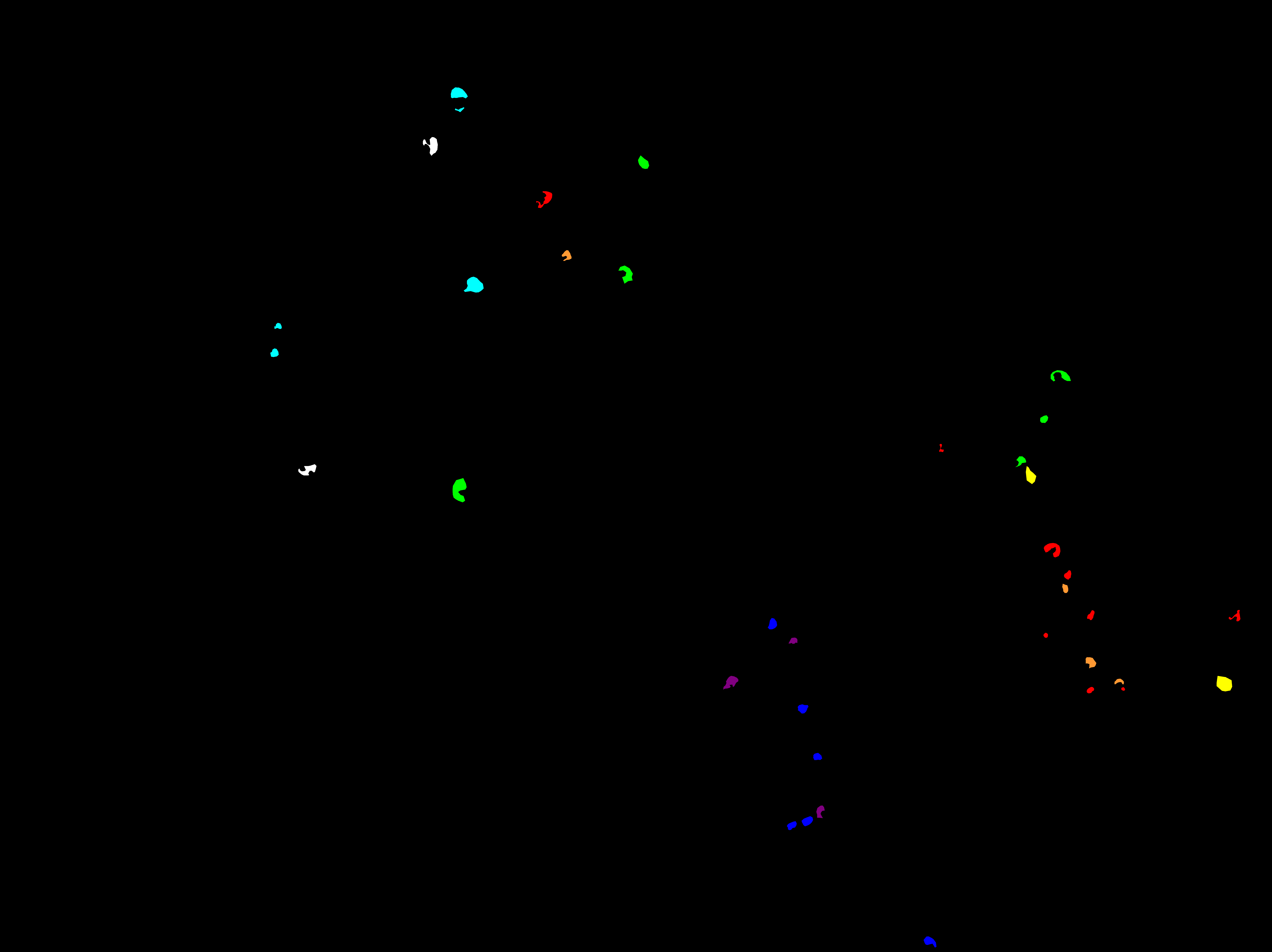}
	}
	\caption{Image of the Serra do Cip{\'o} dataset and the groundtruth.
		Legend -- Black: unclassified/background. Red (train)/Orange (test): {\em Bowdichia virgilioides}. Green (train)/Yellow (test): Collection of evergreen species. Blue (train)/Purple (test): {\em Eremanthus erythropappus}. Cyan (train)/White (test): {\em Vochysia cinnamomea}.
	}
	\label{fig:serracipo}
\end{figure}

Images of this dataset were acquired by a Canon SX260 RGB camera aboard a fixed-wing UAV from October 2015 to February 2017, one for each month (except January and December 2016), totaling \textbf{15 images (or timestamps)}.
The acquired aerial photographs were mosaiced into a single orthoimage, with $6,786\times9,069$ pixels, that was then labeled by experts as belonging to one of the four possible plant species, as shown in Table~\ref{tab:serracipo}.
The imaged vegetation comprises a \emph{campo rupestre} vegetation of the Brazilian Cerrado biome, including a mixture of grasses, shrubs, and trees growing over sandy or rocky substrate.
Specifically, all images were acquired over the Serra do Cip{\'o} region, a mountainous and highly biodiverse and heterogeneous landscape in southern-central Brazil.
Example of a timestamp and the ground truth are presented in Figure~\ref{fig:serracipo}.


\subsection{Baselines} \label{subsec:baselines}


Several techniques were considered as baselines for both datasets:
(i) recurrence plots (RP)~\cite{rp1987,Faria2016PRLb,DBLP:conf/icpr/SouzaSB14}, which are first calculated over the temporal series and then described by hand-crafted descriptors (such as LBP histograms) to be then finally classified by machine learning methods (Linear SVM and Random Forest (with 400 trees), in this case).
(ii) MC-DCNN~\cite{zheng2014time}, a deep-learning-based method that learns features for each and every single channel (or band) and then, concatenates them to be used in the classifier layer, and
(iii) 2-D CNN~\cite{kussul2017deep}, another deep learning approach that performs pixel-classification using a temporal-wise concatenation of image.

For the Itirapina dataset, the method proposed in~\cite{Almeida2014EcoInfo} was also employed as baseline.
In that work, temporal series are described using hand-crafted (color and texture) features, which are provided to a multi-scale classifier (\textbf{MSC}) responsible for performing a late-fusion classification of input images.


\subsection{Experimental Protocol} \label{subsec:protocol}

For the Itirapina dataset, the protocol of~\cite{Almeida2014EcoInfo} was employed.
Particularly, some instances of \textit{Aspidosperma tomentosum} and of \textit{Miconia rubiginosa} were considered for testing.
Pixels of these instances have no influence on the model which is trained using exclusively the remaining samples.
Since this dataset comprises 36 days, the proposed network has 36 branches, one for each day, that receive as input a depth-wise concatenation of all hourly images of that day.

For the Serra do Cip{\'o} dataset, two protocols were used to validate the effectiveness of the proposed method.
The \textbf{first one} follows the same concept of the protocol employed in the Itirapina dataset.
Specifically, all segments (instances) of this dataset were randomly divided into two independent sets: train and test.
The former set is composed of approximately 80\% of the annotated instances while the latter has the remaining 20\%.
Both sets have instances of all classes and the test set has, at least, two segments of each class.
Aside from this, the first protocol considers the whole temporal series as input.
In this case, the proposed network has 15 input branches, one for each timestamp (or month).


The \textbf{second protocol} aims to analyze the robustness of the proposed method to missing data.
Moreover, this protocol is based on the phenology study and its cyclic temporal series.
Precisely, phenological monitoring is cyclic, as plants have a well-defined life cycle along the months.
Therefore, in this protocol, the proposed method is trained and validated using different cycles of the plants (which have missing data).
This process aims at learning and understanding the full life cycle of the plants, creating a model resilient to plant changes over the years.
Specifically, in this protocol, the network is designed to have the dropout layer (with probability $1/12$) and 12 input branches, one for each month, completing the full cycle of a year.
In this case, all images from the largest sequence of time series (i.e., 10 images from 2016, which, compared to a full cycle of the year, has 2 missing data) were used to train the network, while the second largest sequence (i.e., 3 images of 2015, which has 9 missing timestamps) is used for testing.

All results reported are in terms of average accuracy, computed as the average (per-class) ratio of correctly classified samples. 
Thus, it is calculated for each class and, therefore, is independent of any bias connected to class size.
Note that, for both datasets, only annotated pixels are used, while background ones are ignored.
Hence, the results report the average accuracy related only to annotated pixels.

The proposed technique was created using TensorFlow framework\footnote{\url{http://tensorflow.org/} (As of October 2018).} and the code has been made publicly available at~\url{https://github.com/keillernogueira/spatio-temporal-phenological-segmentation/}.
All experiments were performed on a 64 bit Intel i7 5820K machine with 3.3GHz of clock, 64GB of RAM memory and a GeForce GTX Titan X with 12GB of memory under a 9.0 CUDA version. Ubuntu version 16.04.4 LTS was used as operating system.
During training, aforementioned protocols used the following hyper-parameters: learning rate, weight decay, momentum, and number of iterations are 0.01, 0.0005, 0.9, and 200,000, respectively.
After every 50,000 iterations, the learning rate is reduced using the exponential decay\footnote{\url{https://www.tensorflow.org/api_docs/python/tf/train/exponential_decay} (As of October 2018).}.	

\section{Experimental Results} \label{sec:results}


\subsection{Time Series Classification} \label{subsec:spatio-temporal}

For the Itirapina dataset, obtained results are presented in Table~\ref{tab:itaparina_results}.
The worst results were yielded by the methods based on RP-based technique~\cite{Faria2016PRLb}.
Such approaches take only a few hours to be trained but are outperformed by all other approaches.
This may be justified by the fact that those techniques are based on hand-crafted features which are used to train a single model.
This is the same reason why the MSC~\cite{Almeida2014EcoInfo} method, which achieves good results (76.00\% and 90.00\% of accuracy for {\em Aspidosperma} and {\em Rubiginosa} classes, respectively), is outperformed by the ConvNets.
Precisely, this method, that requires approximately 8 hours of training to create an ensemble of models, also does not have feature learning but only combines hand-crafted features.
Among the deep learning-based methods, MC-DCNN~\cite{zheng2014time} achieved great results (96.74\% and 99.17\% of accuracy for {\em Aspidosperma} and {\em Rubiginosa} classes, respectively) taking only 12 hours to train.
However, all these techniques were outperformed by the 2-D CNN~\cite{kussul2017deep} and the proposed approach, which achieved the best result for both classes (100.00\%), taking 24 and 16 hours to train, respectively.

\begin{table}[]
	\centering
	\caption{Results for the Itirapina dataset.}
	\label{tab:itaparina_results}
	\resizebox{\columnwidth}{!}{ 
		\begin{tabular}{@{}lrrr@{}}
			\toprule
			\multicolumn{1}{c}{\multirow{2}{*}{\textbf{Method}}}                       & \multicolumn{1}{c}{\multirow{2}{*}{\textbf{\begin{tabular}[c]{@{}c@{}}Training\\ Time (h)\end{tabular}}}} & \multicolumn{2}{c}{\textbf{Accuracy (\%)}}                                          \\ \cmidrule(l){3-4} 
			\multicolumn{1}{c}{}                                                       & \multicolumn{1}{c}{}                                                                                      & \multicolumn{1}{c}{\textbf{Aspidosperma}} & \multicolumn{1}{c}{\textbf{Rubiginosa}} \\ \midrule
			\textbf{Recurrence Plots~\cite{Faria2016PRLb} + SVM} & 3                                                                                                         & 64.35                                     & 73.15                                   \\
			\textbf{Recurrence Plots~\cite{Faria2016PRLb} + RF}  & 3                                                                                                         & 65.14                                     & 86.01                                   \\
			\textbf{MSC~\cite{Almeida2014EcoInfo}}               & 8                                                                                                         & 76.00                                     & 90.00                                   \\
			\textbf{MC-DCNN~\cite{zheng2014time}}                & 12                                                                                                         & 96.74                                     & 99.17                                   \\
			\textbf{2-D CNN~\cite{kussul2017deep}}               & 24                                                                                                        & \textbf{100.00}                           & \textbf{100.00}                         \\
			\textbf{Multi-temporal ConvNet (ours)}                                     & 16                                                                                                        & \textbf{100.00}                           & \textbf{100.00}                         \\ \bottomrule
		\end{tabular}
	}
\end{table}

For the Serra do Cip{\'o} dataset, results, based on the first protocol, are presented in the first part of Table~\ref{tab:serra_results}.
Again, the worst results were produced by the methods based on Recurrence Plots technique~\cite{Faria2016PRLb}.
Although achieving the worst accuracy such methods are the fastest to train, taking only 2 hours.
The MC-DCNN~\cite{zheng2014time} method, which requires 3 hours to train a new model, achieved great results (97.83\%).
However, the best outcomes were yielded by the 2-D CNN~\cite{kussul2017deep} and the proposed method, which achieved almost perfect classification (around 99\%) taking approximately 4,5 hours. 

Two interesting facts can be pointed out based on the obtained results:
(i) 2-D CNN~\cite{kussul2017deep} and the proposed approach achieved very similar results and should be better analyzed, and
(ii) such methods produced (almost) perfect classifications, which clearly motivates an ablation study to analyze if all timestamps are necessary.
Both issues are better analyzed next.


\begin{table}[]
	\centering
	\caption{Results for the Serra do Cip{\'o} dataset.}
	\label{tab:serra_results}
	\resizebox{\columnwidth}{!}{
		\begin{tabular}{@{}clrr@{}}
			\toprule
			\textbf{Protocol}                    & \multicolumn{1}{c}{\textbf{Method}}    & \multicolumn{1}{c}{\textbf{\begin{tabular}[c]{@{}c@{}}Training\\ Time\end{tabular}}} & \multicolumn{1}{c}{\textbf{Accuracy (\%)}} \\ \midrule
			\multirow{5}{*}{\textbf{Protocol 1}} & \textbf{Recurrence Plots~\cite{Faria2016PRLb} + SVM}        & 2                                                                                    & 57.12                           \\
			& \textbf{Recurrence Plots~\cite{Faria2016PRLb} + RF}         & 2                                                                                    & 60.96                           \\
			& \textbf{MC-DCNN~\cite{zheng2014time}}                       & 3                                                                                    & 97.83                           \\
			& \textbf{2-D CNN~\cite{kussul2017deep}}                       & 4                                                                                    & 99.33                 \\
			& \textbf{Multi-temporal ConvNet (ours)} & 5                                                                                   & \textbf{99.78}                 \\ \midrule
			\multirow{5}{*}{\textbf{Protocol 2}} & \textbf{Recurrence Plots~\cite{Faria2016PRLb} + SVM}        &                                                                                      & 26.39                                      \\
			& \textbf{Recurrence Plots~\cite{Faria2016PRLb} + RF}         &                                                                                      & 27.96                                      \\
			& \textbf{MC-DCNN~\cite{zheng2014time}}                       &                                                                                      & 34.11                                      \\
			& \textbf{2-D CNN~\cite{kussul2017deep}}                       &                                                                                      & 35.24                                      \\
			& \textbf{Multi-temporal ConvNet (ours)} &                                                                                      & \textbf{50.70}                             \\ \bottomrule
		\end{tabular}
	}
\end{table}

\subsection{Ablation Study} \label{subsec:ablation}

An ablation study was conducted to assess the robustness of methods that produced perfect classification.
It can be divided into three steps: 
(i) for each analyzed method, a model is trained for each timestamp,
(ii) a correlation analysis (based on~\cite{kuncheva2003measures}) is performed for all trained models, and
(iii) the most suitable images (timestamps) (with the lowest correlation) are then combined and used to train new models.
Note that the combination of the most suitable images is performed from the smallest possible time series (with only two images) to the largest (the whole time series, as in the previous section).

For the Itirapina dataset, as presented in Table~\ref{tab:itaparina_ablation}, the best result of the 2-D CNN~\cite{kussul2017deep} was achieved using the image of September 10, 2011.
However, the proposed method achieved a better result using the same image for the {\em Aspidosperma} class, whereas reproducing the same perfect classification for the {\em Rubiginosa} class.
The same outcome is generated analyzing the models trained with the proposed approach.
In this case, the method achieved perfect classification for both classes using only a timestamp (August 30, 2011).
The 2-D CNN~\cite{kussul2017deep} trained using the same image achieved perfect classification for the {\em Rubiginosa} class but worse for the {\em Aspidosperma} one.
This shows the effectiveness of the proposed method even when using only one timestamp.

\begin{table}[]
	\caption{Results for the ablation study over the Itirapina dataset.}
	\label{tab:itaparina_ablation}
	\resizebox{\columnwidth}{!}{ 
		\begin{tabular}{@{}clrr@{}}
			\toprule
			\multirow{2}{*}{\textbf{\begin{tabular}[c]{@{}c@{}}Time\\ Series\end{tabular}}} & \multicolumn{1}{c}{\multirow{2}{*}{\textbf{Method}}}         & \multicolumn{2}{c}{\textbf{Accuracy (\%)}}                                          \\ \cmidrule(l){3-4} 
			& \multicolumn{1}{c}{}                                         & \multicolumn{1}{c}{\textbf{Aspidosperma}} & \multicolumn{1}{c}{\textbf{Rubiginosa}} \\ \midrule
			\multirow{2}{*}{\textbf{Sept, 10 2011}}                                          & \textbf{2-D CNN~\cite{kussul2017deep}} & 94.72                                     & \textbf{100.00}                         \\
			& \textbf{Multi-temporal ConvNet (ours)}                       & \textbf{100.00}                           & \textbf{100.00}                         \\ \midrule
			\multirow{2}{*}{\textbf{Aug, 30 2011}}                                          & \textbf{2-D CNN~\cite{kussul2017deep}} & 59.55                                     & \textbf{100.00}                         \\
			& \textbf{Multi-temporal ConvNet (ours)}                       & \textbf{100.00}                           & \textbf{100.00}                         \\ \bottomrule
		\end{tabular}
	}
\end{table}

As presented in Table~\ref{tab:serra_ablation}, for the Serra do Cip{\'o} dataset, the 2-D CNN~\cite{kussul2017deep} and the proposed approach achieved almost perfect segmentation using 6 and 4 timestamps, respectively.
Moreover, the results of the 2-D CNN~\cite{kussul2017deep} (99.26\%) is slightly outperformed by the proposed approach (99.55\%) when using 6 timestamps.
On the other hand, the 2-D CNN~\cite{kussul2017deep}, using only 4 timestamps, is not able to produce satisfactory results (97.41\%), like the ones produced by the proposed technique (99.51\%).
Such results validate previous conclusions that the proposed method can learn a discriminative representation.

\begin{table}[]
	\caption{Results for the ablation study over the Serra do Cip{\'o} dataset.}
	\label{tab:serra_ablation}
	\resizebox{\columnwidth}{!}{ 
		\begin{tabular}{@{}clr@{}}
			\toprule
			\textbf{\begin{tabular}[c]{@{}c@{}}Time\\ Series\end{tabular}}                          & \multicolumn{1}{c}{\textbf{Method}}                          & \multicolumn{1}{c}{\textbf{Accuracy (\%)}} \\ \midrule
			\multirow{2}{*}{\textbf{\begin{tabular}[c]{@{}c@{}}Oct, 2015; Feb, 2016; Jun, 2016;\\Aug, 2016; Oct, 2016; Jan, 2017\end{tabular}}} & \textbf{2-D CNN~\cite{kussul2017deep}} & 99.26                           \\
			& \textbf{Multi-temporal ConvNet (ours)}                       & \textbf{99.55}                 \\ 
			\midrule
			\multirow{2}{*}{\textbf{\begin{tabular}[c]{@{}c@{}}Oct, 2015; Feb, 2016;\\ Oct, 2016; Jan, 2017\end{tabular}}} & \textbf{2-D CNN~\cite{kussul2017deep}} & 97.41                           \\
			& \textbf{Multi-temporal ConvNet (ours)}                       & \textbf{99.51}                 \\ \bottomrule
		\end{tabular}
	}
\end{table}

\subsection{Missing Data} \label{subsec:missing}


To analyze the robustness of the proposed method to missing data, we conducted experiments employing the second protocol, proposed in Section~\ref{subsec:protocol}, which based on the life cycle of the plants.
To allow the evaluation of the baselines, we proposed an adaptation, based on dropout method~\cite{srivastava2014dropout}, just as in the proposed method.
All approaches were trained considering a year cycle aiming at learning and understanding the full life cycle of the plants.
Therefore, all 10 images captured in 2016 were used to train the network, while the 3 images from 2015 were employed for testing.
Results are presented in the second part of Table~\ref{tab:serra_results}.
Once again, the worst result was yielded by the Recurrence Plots technique~\cite{Faria2016PRLb} (around 27\%).
Baselines based on ConvNets outperformed such technique achieving very similar performance (around 35\%).
However, all baselines were outperformed by the proposed method, that yielded 50.70\% of average accuracy.
This result shows the ability of the proposed method in:
(i) learning a spatio-temporal representation to perform vegetation pixel-classification while dealing with missing data, and
(ii) capturing the cycles of plants while being resilient to their changes over the year.

\section{Conclusion} \label{sec:conclusion}

We proposed a novel convolutional network to perform spatio-temporal pixel-classification over high-resolution vegetation images.
Specifically, the network has two parts:
the first one is composed of branches (in which images of each timestamp are independently processed) responsible for extracting spatial information from each entry in the time series, while the second one receives and combines all previous information extracted by the branches generating a final spatio-temporal representation.
Experimental results showed the effectiveness of the proposed method to perform spatio-temporal pixel-classification.
In fact, an ablation study showed that the proposed technique achieved state-of-the-art performance, in terms of average accuracy, in two temporal datasets of plant species outperforming traditional and deep learning-based baselines.
Furthermore, experiments showed that the proposed approach has support to time series with missing data, a common aspect in temporal datasets.
As future work, we intend to better evaluate the robustness of the proposed method to deal with missing data, and analyze the proposed method in other applications.



\bibliographystyle{IEEEtran}
\bibliography{bibliography}

\end{document}